\documentclass{article}

\PassOptionsToPackage{numbers, compress}{natbib}

\usepackage[preprint]{neurips_2019}

\usepackage{natbib}
\usepackage[utf8]{inputenc} 
\usepackage[T1]{fontenc}    
\usepackage{hyperref}       
\usepackage{url}            
\usepackage{booktabs}       
\usepackage{amsfonts}       
\usepackage{nicefrac}       
\usepackage{microtype}      
\usepackage{amsfonts,amssymb}
\usepackage{graphicx}
\usepackage{subfig}
\usepackage{multirow}

\title{HGC: Hierarchical Group Convolution for Highly Efficient Neural Network}

\author{
  Xukai Xie \\
  Tianjin university \\
  \texttt{xkxie@tju.edu.cn} \\
  \And
  Yuan Zhou\textsuperscript{*}\\
  Tianjin university \\
  \texttt{zhouyuan@tju.edu.cn} \\
  \And
  Sun-Yuan Kung \\
  Princeton University \\
  \texttt{kung@princeton.edu} \\
}

\begin{document}

\maketitle

\begin{abstract}
  Group convolution works well with many deep convolutional neural networks (CNNs) that can effectively compress the model by reducing the number of parameters and computational cost. Using this operation, feature maps of different group cannot communicate, which restricts their representation capability. To address this issue, in this work, we propose a novel operation named Hierarchical Group Convolution (HGC) for creating computationally efficient neural networks. Different from standard group convolution which blocks the inter-group information exchange and induces the severe performance degradation, HGC can hierarchically fuse the feature maps from each group and leverage the inter-group information effectively. Taking advantage of the proposed method, we introduce a family of compact networks called HGCNets. Compared to networks using standard group convolution,  HGCNets have a huge improvement in accuracy at the same model size and complexity level. Extensive experimental results on the CIFAR dataset demonstrate that HGCNets obtain significant reduction of parameters and computational cost to achieve comparable performance over the prior CNN architectures designed for mobile devices such as MobileNet and ShuffleNet.

\end{abstract}

\section{Introduction}
\label{I}
Deep convolutional neural networks (CNNs) have shown remarkable performance in many computer vision tasks in recent years. In order to achieve higher accuracy for major tasks such as image classification, building deeper and wider CNNs \cite{2,3,4,11} is the primary trend. However, deeper and wider CNNs usually have hundreds of layers and thousands of channels, which come with an increasing amount of parameters and computational cost. For example, one of the classic networks, VGG16 \cite{2} with 130 million parameters needs more than 30 billion floating-point operations (FLOPs) to classify a single image, it fails to achieve real-time classification even with a powerful GPU. And many real-world applications often need to be performed on limited resource in real-time, e.g., mobile devices. Thereby, the model should be compact to reduce computational cost and achieve better trade-off between efficiency and accuracy.
\begin{figure*}[htbp]\label{figure1}
\label{XVI}
\centering
\subfloat[]{\includegraphics[width = 0.26\columnwidth]{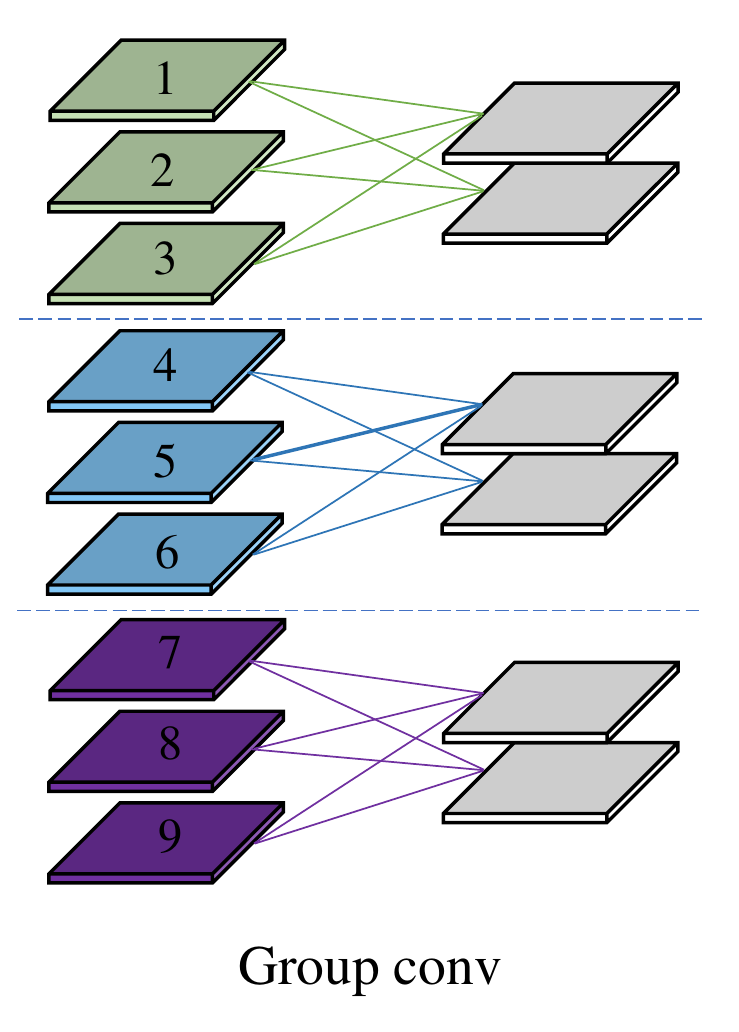}}\
\hspace{1cm}
\label{fig1b}
\subfloat[]{\includegraphics[width = 0.44\columnwidth]{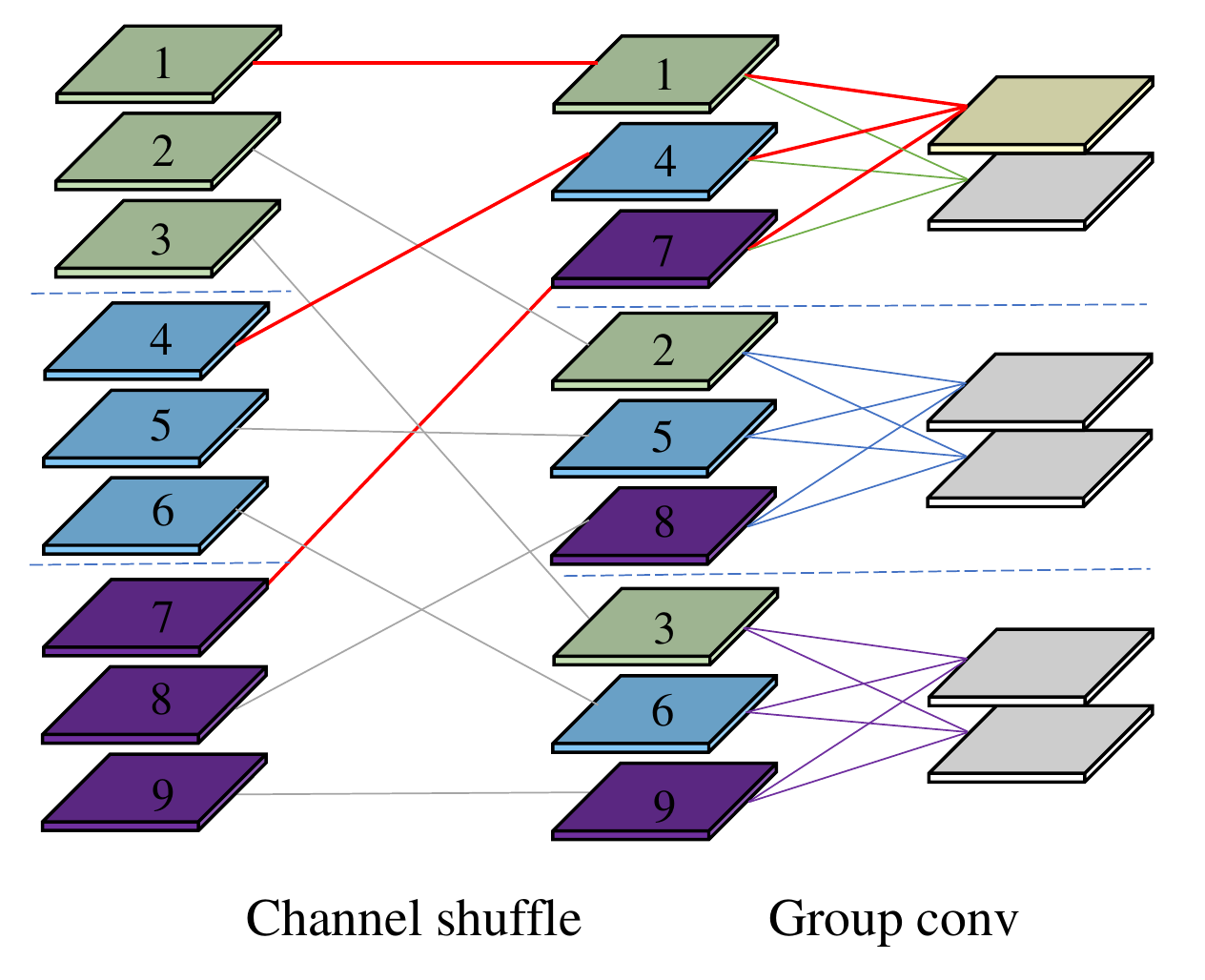}}\
\caption{ An illustration of group convolutions. (a) is a standard group convolution, which severely blocks the information flow between channels of each group. (b) is a group convolution with shuffle operation to facilitate inter-group information exchange, but still suffers from much loss of inter-group information. Red lines show that each output only relates to three input channels.
}
\end{figure*}

Recently, many research work focus on the field of model compression \cite{5,6,7,8,10}. These works can be separated into two main kinds of approaches: compression for per-trained network and efficient architecture design. The compressing approach usually bases on traditional compression techniques such as pruning and quantization which removes connections to eliminate redundancy or reduce the number of bits to represent the parameters. These approaches are simple and intuitive, but always needs multiple steps, i.e., pretraining and compressing, thus cannot do an end-to-end training at one time. The second approach trains model from scratch in a fully end-to-end manner. It usually utilizes a sequence of sparsely-connected convolutions rather than the standard fully-connected convolution to design new efficient architectures. For instance, in the ShuffleNet \cite{8}, the original $3\times3$ convolution is replaced with a $3\times3$ depthwise convolution, while the $1\times1$ convolution is substituted with a pointwise group convolution. The application of group convolution significantly reduces amount of parameters and the computational cost. However, the group convolution blocks the information flow between channels of each group, as shown in Figure 1(a), the $G$ groups are computed independently from completely separate groups of input feature maps, thus there is no interaction between each group and leads to severe performance degradation. Although ShuffleNet introduces a channel shuffle operation to facilitate inter-group information exchange, it still suffers from the loss of inter-group information. As shown in Figure 1(b), even with a shuffle operation, a large portion of the inter-group information cannot be leveraged. This problem is aggravated when number of channel groups increases.

To solve the above issue, we propose a novel operation named Hierarchical Group Convolution (HGC) to effectively facilitate the interaction of information between different groups. In contrast to common group convolution, HGC can hierarchically fuse the feature maps from each group and leverage the inter-group information effectively. Specifically, we split the input feature maps of a layer into multiple groups, in which the first group features are extracted by a group of filters; output feature maps of the previous group are concatenated with the next group of input feature maps, and then feed to the next group of filters. This process repeats until all input feature maps are included. By exploiting the HGC operation and depthwise seperable convolution, we introduce the HGC module, a powerful and effective unit to build a highly efficient architecture called HGCNets. A series of controlled experiments show the effectiveness of our design. Compared to other structures, HGCNets perform better in alleviating the loss of inter-group information, and thus achieve substantial improvement as the group number increases.

Our work brings following contributions and benefits: First, a new hierarchical group convolution operation is proposed to facilitate the interaction of information between different groups of feature maps and leverage the inter-group information effectively. Second, our proposed HGCNets achieve higher classification accuracy than prior compact CNNs at the same or even lower complexity.

The rest of this paper is organized as follow: Section \ref{II} provides an overview of the related work on model compression. The details of the proposed Hierarchical Group Convolution operation is introduced in Section \ref{III}. In Section \ref{IV}, we describe the structure of the HGC module and HGCNets architecture. The performance evaluation of the proposed method is described in Section \ref{V}. Finally, we conclude this paper in Section \ref{VI}.

\section{Related work}
\label{II}
We first review two main approaches of model compression: compression for pre-trained networks and designing efficient architectures, which inspire our work. Next, we review the group convolution that form the basis for HGCNets.

\subsection{Compression for pre-trained networks}
Most of works applied this approach improve the efficiency of CNNs via weight pruning \cite{12,13,19} and quantization \cite{14}. These approaches are effective because deep neural networks often have a substantial number of redundant weights that can be pruned or quantized without sacrificing much accuracy. For convolutional neural networks, different pruning techniques may lead to different levels of granularity. Fine-grained pruning, e.g., independent weight pruning \cite{12}, generally achieves a high degree of sparsity. However, it requires storing a large number of indices, and relies on special hardware/software accelerators. In contrast, coarse-grained pruning methods such as filter-level pruning \cite{19} achieve a lower degree of sparsity, but the resulting networks are much more regular, which facilitates efficient implementations. These approaches are simple and intuitive, however, iterative optimization strategy is commonly utilized in these approaches, which slows down the training procedure.

\subsection{Designing efficient architectures}
Considering the above-mentioned limitations, some researchers go other way to directly design efficient network architectures \cite{7,8,20} that can be trained end-to-end using smaller filters, such as depthwise separable convolution, group convolution, and etc. Two well-known applicants of this kind of approach that are sufficiently efficient to be deployed on mobile devices are MobileNet \cite{7} and ShuffleNet \cite{8}. MobileNet exploited depthwise separable convolution as its building unit, which decompose a standard convolution into a combination of a depthwise convolution and a pointwise convolution. ShuffleNet utilize depthwise convolution and pointwise group convolution into the bottleneck unit, and proposed the channel shuffle operation to enable inter-group information exchange. Compact networks can be trained from the scratch, so the training procedure is very fast. Moreover, the model can be further compressed combined with the aforementioned model compression methods which are orthogonal to this approach, e.g., Huang \cite{18} combined the channel pruning and group convolution to sparsify networks, however, this channel pruning methods obtain a sparse network based on a complex training procedure that requires significant cost of offline training and directly removing the input feature maps typically has limited compression and speedup with significant accuracy drop.In addition to the methods described above, some other approaches such as low-rank factorization \cite{15} and knowledge distillation \cite{16} can also efficiently accelerate deep neural network.
\subsection{Group Convolution}
Group convolution is a special case of a sparsely connected convolution. It was first used in the AlexNet \cite{1} architecture, and has more recently been popularized by their successful application in ResNeXt \cite{17}. Standard convolutional layers generate $O$ output feature maps by applying convolutional filters over all $I$ input feature maps, leading to a computational cost of $I\times O$. In comparison, group convolution reduces this computational cost by partitioning the input features into G mutually exclusive groups and each group produces its own outputs—reducing the computational cost by a factor $G$ to $\frac{I\times O}{G}$ . However, the grouping operation usually compromises performance because there is no interaction among groups. As a result, information of feature maps in different groups is not combined, as opposed to the original convolution that combines information of all input channels, which restricts their representation capability. To solve this problem, in ShuffleNet \cite{8}, a channel shuffle operation is proposed to permute the output channels of group convolution and makes the output better related to the input. But any output group still only accesses $\frac{I}{G}$  input feature maps and thus collects partial information. Due to this reason, ShuffleNet has to employ a deeper architecture than MobileNet to achieve competitive results.

\section{Hierarchical Group Convolution}
\label{III}
\subsection{Motivation}
\label{XIII}
In modern deep neural networks, the size of convolutional filters is mostly $3\times 3$ or $1\times 1$, and the main computational cost is from the convolutional layer, that the fully connected layer can be considered as a special case of the $1\times 1$ convolutional layer. To reduce the parameters in convolution operation, an extremely efficient scheme is to replace standard $3\times 3$ convolution by a $3\times 3$ depth-wise separable convolution \cite{27} followed by interleaved $1\times 1$ group convolution \cite{20,8}. This scheme significantly reduces the model size and therefore attracts increasing attention.

Since the $1\times 1$ filters are non-seperable, group convolution becomes a hopeful and feasible solution and works well with many deep neural network architectures. However, preliminary experiments show that a naive adaptation of group convolution in the $1\times 1$ convolutional layer leads to drastic reductions in accuracy especially in dense architectures. As analyzed in CondenseNet \cite{18}, this is caused by the fact that the inputs to the $1\times 1$ convolutional layer are concatenations of feature maps generated by preceding layers and they have an intrinsic order or they are far more diverse. The hard assignment of these features to disjoint groups hinders effective feature reuse in the network. More specifically, as investigated in network explanation, individual feature maps across different layers play different roles in the network, e.g., features from shallow layers usually encode low-level spatial visual information like edges, corners, circles, etc., and features from deep layers encode high-level semantic information. Group convolution severely blocks the inter-group information exchange and induce the severe performance degradation. In order to facilitate the fusion of feature maps from each group and leverage the inter-group information effectively, we develop a novel approach, named hierarchical group convolution operation that efficiently overcomes the side effects brought by the group convolution.
\begin{figure}
\centerline{\includegraphics[width = 0.6\columnwidth]{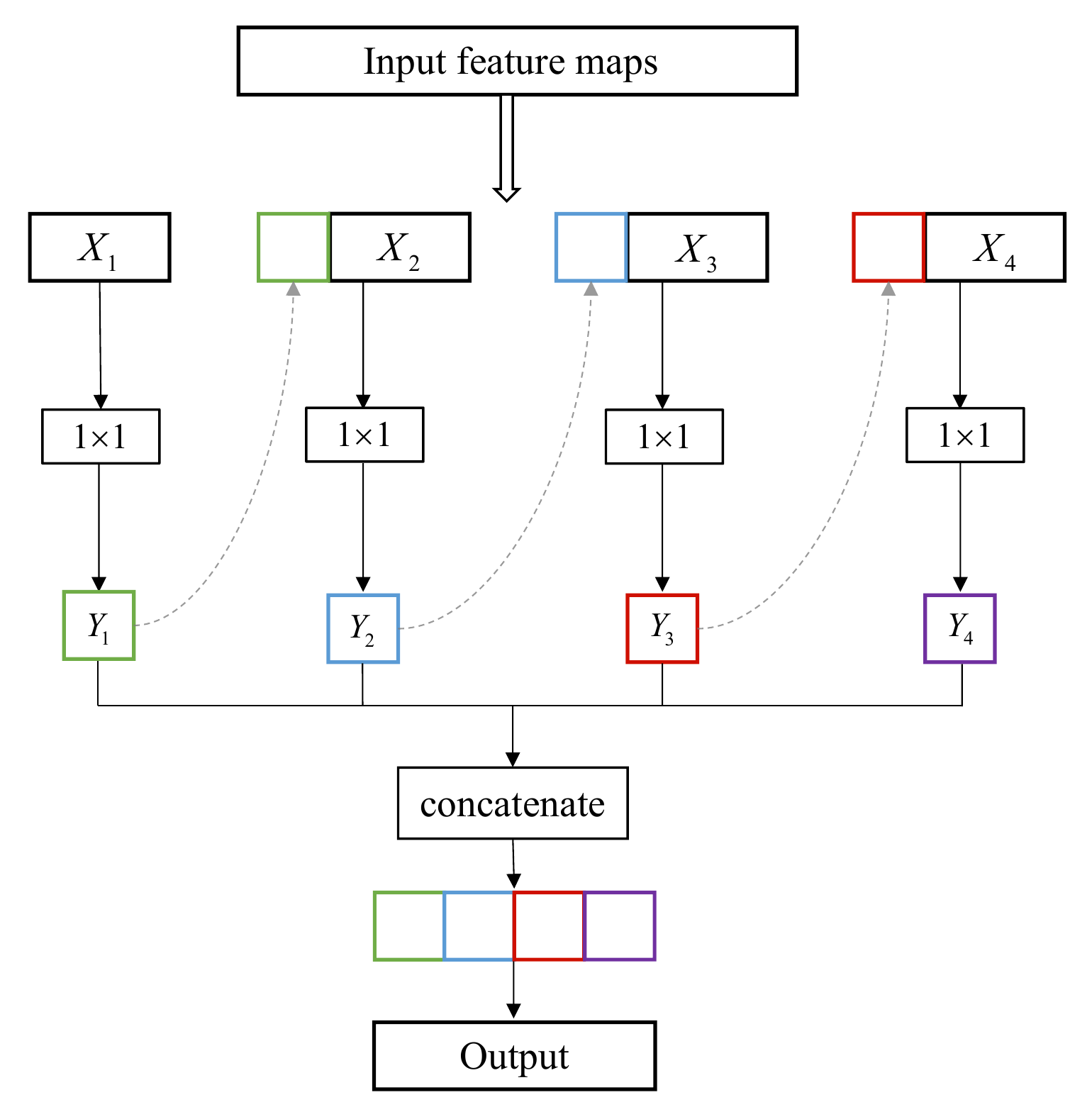}}
\caption{The proposed Hierarchical Group Convolution operation.}
\label{fig2}
\end{figure}

\subsection{Details of Hierarchical Group Convolution}
\label{X}
Details of the proposed Hierarchical Group Convolution are shown in Figure 2. Generally, a $1\times 1$ standard convolutional layer transforms the input feature maps $X\in \mathbb{R}^{I\times H_{in}\times W_{in}}$ into the output feature maps $Y\in \mathbb{R}^{O\times H_{out}\times W_{out}}$ by using the filters $W\in \mathbb{R}^{O\times I\times 1\times 1}$. Here, $I$ and $O$ is the number of the input feature maps and the output feature maps respectively. In HGC operation, the input channels and filters are divided into $G$ groups respectively, i.e., $\frac{I}{G}$ input channels and $\frac{O}{G}$ filters in each group, denote as $X=\left\{ X_1, X_2, \cdots,X_G \right\}$ where each $X_i\in\mathbb{R}^{\frac{I}{G}\times H_{in}\times W_{in}}$, and $W=\left\{ W_1, W_2, \cdots,W_G \right\}$, where $W_i\in\mathbb{R}^{\frac{O}{G}\times \frac{I}{G}\times 1\times 1}$  when $i=1$, and $W_i\in\mathbb{R}^{\frac{O}{G}\times (\frac{O}{G}+\frac{I}{G})\times 1\times 1}$ when $i\in \left\{ 2, 3, \cdots,G \right\}$. Except that the first group feature maps directly go through the $W_1$, the feature group$X_i$ is concatenated with the output $Y_{i-1}$ on the channel dimension, and then fed into $W_i$. Thus, the $Y_i$ can be formulated as follows:

\begin{equation}
\label{1}
    Y_i=\left\{
    \begin{array}{rcl}
    &X_i*W_i   & {i = 1}\\
    &concatenate(X_i,Y_{i-1})*W_i   & {1<i\leq G}\\
    \end{array}
\right.
\end{equation}

where * represents the $1\times 1$ convolutional operation. For simplicity, the biases are omitted for easy presentation. After all input feature maps are processed, we finally concatenate each $Y_i$ as the output of HGC.

Notice that each $1\times 1$ convolutional operator could potentially receive information from all feature subsets $\left\{ X_k, k\leq i\right\}$ of the previous layer. Each time a feature group $X_k$ go through a $1\times 1$ convolutional operator, the output result can have more information from input feature maps. The split and concatenation strategy can effectively process feature maps with less parameters. The parameters of the HGC is calculated as bellow:
\begin{equation}
\label{2}
\frac{O}{G}\times \frac{I}{G}\times 1\times 1+(G-1)\times \frac{O}{G}\times (\frac{O}{G}+\frac{I}{G})\times 1\times 1
\end{equation}

compared with the parameters of standard convolution, the compression ratio $r$ of each layer is:
\begin{equation}
\label{3}
r=(\frac{O}{I\times G}+\frac{1}{G})\times (1-\frac{1}{G})+\frac{1}{G^2} \approx \frac{2}{G}\times (1-\frac{1}{G})+\frac{1}{G^2}
\end{equation}

As can be observed in Eq. 3, HGC contains about $\frac{2}{G}$ fewer parameters than standard convolution. Although with negligible parameters increase than standard group convolution, HGC has stronger ability of feature representation. As will be shown in Section \ref{VII}, HGC has a substantial improvement in accuracy especially in the case of large number of groups.

\section{HGCNet}
\label{IV}
\subsection{HGC module}
\label{XII}
Taking advantage of the proposed HGC operation, we propose a novel HGC module specially designed for efficient neural networks. The HGC module is shown in Figure 3(b). The typical bottleneck structure shown in Figure 3(a) is a basic building block in many modern backbone CNNs architectures, e.g., Densenet \cite{11}. Instead of directly extracting features using a group of $1\times 1$ convolutional filters as in the bottleneck, we use HGC operation with stronger inter-group information exchange ability, while maintaining similar computational load. A channel shuffle operation before the HGC allows for more inter-group information exchange. Finally, feature maps from all groups are concatenated and sent to a computational economical $3\times3$ depthwise seperable convolution to capture spatial information. The usage of batch normalization \cite{9} and nonlinearity \cite{21} is similar to Xception \cite{27}, that we do not use ReLU before depthwise convolution.

As discussed in Section \ref{XIII}, the information contained in each output group gradually increase, which results in each channel has different contribution to latter layers. Thus, we can integrate the SE \cite{28} block to the HGC module to adaptively re-calibrates channel-wise feature responses by explicitly modeling importance of each channel. Our HGC module can benefit from the integration of the SE block, which we have experimentally demonstrated in Section \ref{IX}.
\begin{figure*}[htbp]\label{figure3}
\centering
\subfloat[]{\includegraphics[width = 0.18\columnwidth]{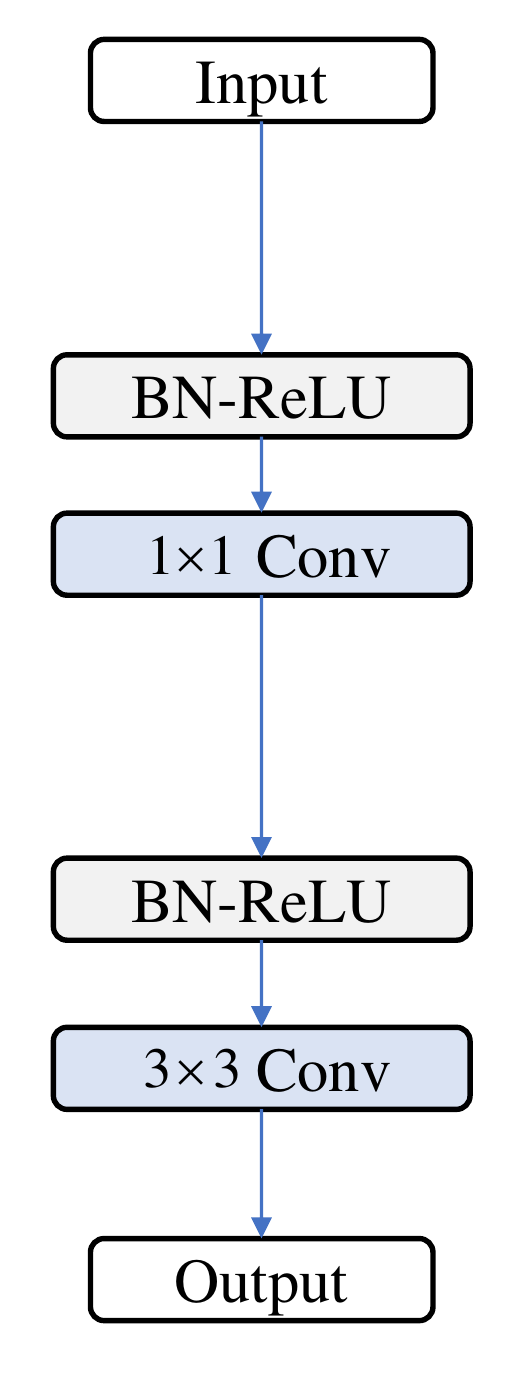}}\
\hspace{1cm}
\subfloat[]{\includegraphics[width = 0.19\columnwidth]{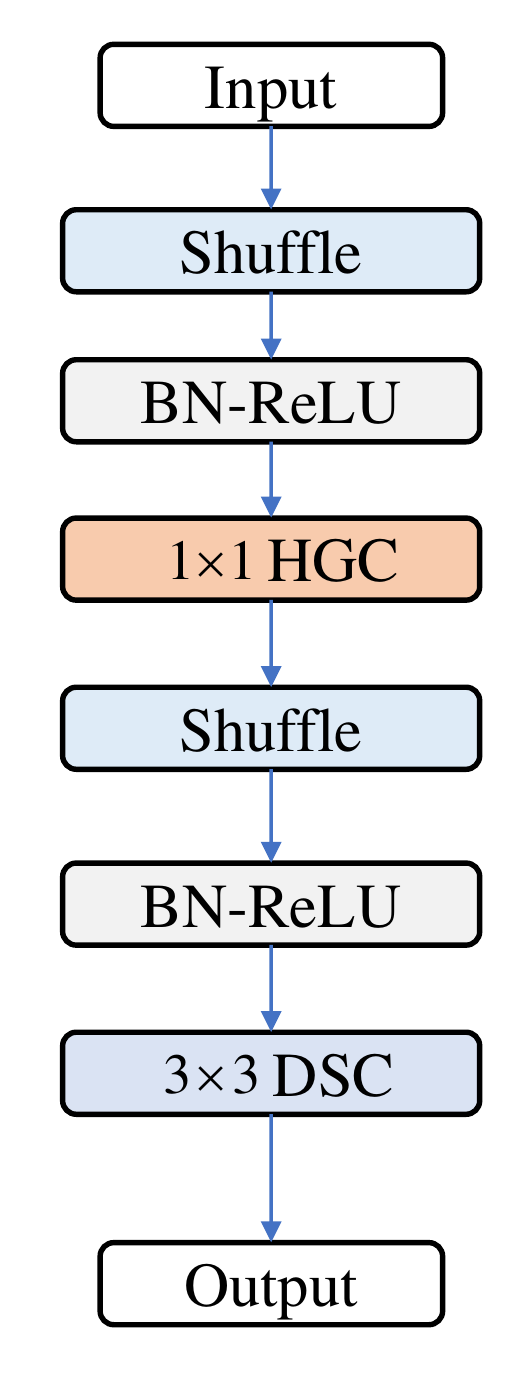}}\
\caption{ Comparison between the bottleneck block and the proposed HGC module. (a)is the typical bottleneck. (b)is the proposed HGC module}
\end{figure*}
\subsection{HGCNet Architecture}
Combined with the efficient HGC module and dense connectivity, we propose HGCNets, a new family of compact neural networks. Similar to CondenseNet \cite{18}, we exponentially increasing the growth rate as the depth grows to increase the proportion of features coming from later layers relative to those from earlier layers due to the fact that deeper layers in DenseNet tend to rely on high-level features more than on low-level features. For simplicity, we multiply the growth rate by a power of 2. The overall architecture of HGCNets for CIFAR classification is group into three stages. The number of HGC module output channels is kept the same to the growth-rate in each stage, and doubled in the next stage.

\section{Experiments}
\label{V}
In this section, we evaluate the effectiveness of our proposed HGCNets on the CIFAR-10, CIFAR-100 \cite{22} image classification datasets. We implement all the proposed models using the Pytorch framework \cite{26}.

\textbf{Datasets}. The CIFAR-10 and CIFAR-100 datasets consist of RGB images of size $32\times 32$ pixels, corresponding to 10 and 100 classes, respectively. Both datasets contain 50,000 training images and 10,000 test images. We use a standard data-augmentation scheme \cite{23,24,25}, in which the images are zero-padded with 4 pixels on each side, randomly cropped to produce $32\times 32$ images, and horizontally mirrored with probability 0.5.
\begin{table}[hbp]
  \caption{Top-1 test error on the CIFAR-10 dataset}
  \begin{center}
  \begin{tabular}{c c c c}
  \toprule
    Model  & FLOPs & Params & Top-1 err. (\%) \\
    \midrule
    SGCNet-42 (G = 1) & 91M & 0.55M & 5.94  \\
    \midrule
    HGCNet-42 (G = 2) & 74M & 0.41M & 6.21 \\
    \midrule
    SGCNet-42 (G = 2) & 62M & 0.33M & 6.46 \\
    \midrule
    HGCNet-42 (G = 4) & 56M & 0.28M & 6.36 \\
    \midrule
    SGCNet-42 (G = 4) & 47M & 0.22M & 6.64 \\
    \midrule
    HGCNet-42 (G = 6) & 49M & 0.23M & 6.52 \\
    \midrule
    SGCNet-42 (G = 6) & 42M & 0.19M & 6.94 \\
  \bottomrule
  \end{tabular}
  \end{center}
\end{table}
\begin{figure*}[htbp]\label{figure4}
\centering
\subfloat[]{\includegraphics[width = 0.47\columnwidth]{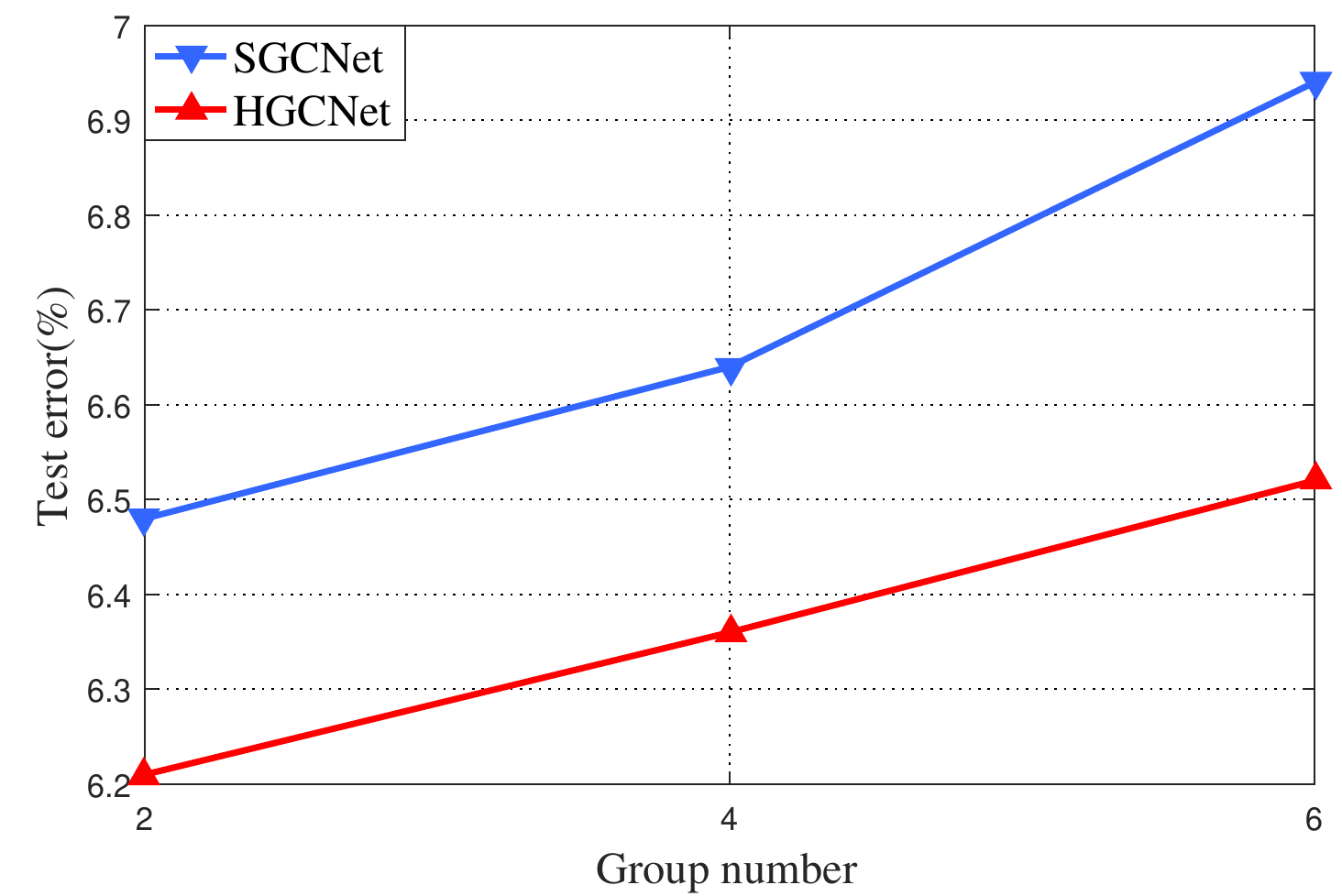}}\
\hspace{0.5cm}
\subfloat[]{\includegraphics[width = 0.48\columnwidth]{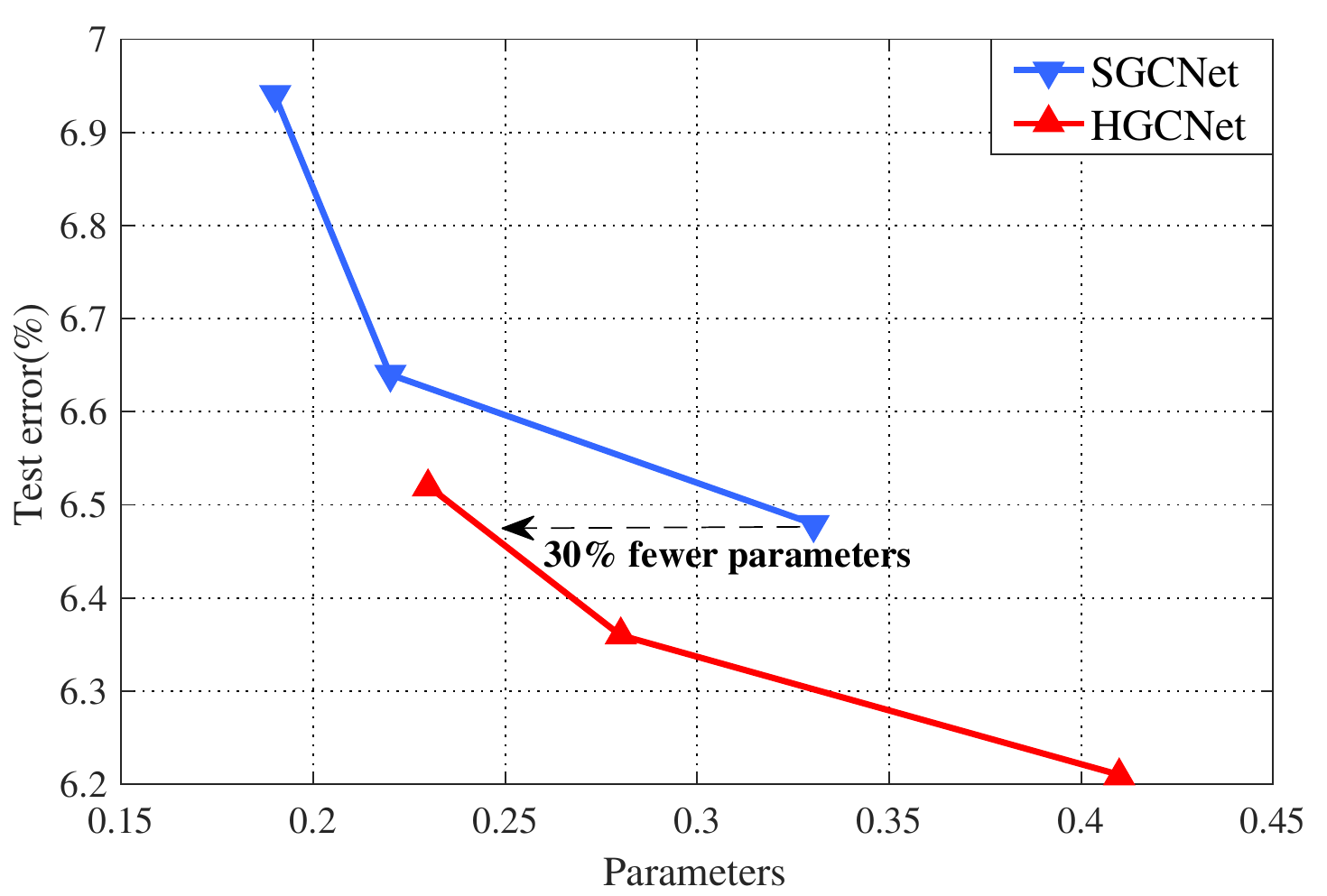}}\
\caption{ Comparison between the SGC and the proposed HGC operation. (a) is the test error of each network under the same number of groups. (b) shows relation between the efficiency and accuracy.}
\end{figure*}
\begin{figure*}[htbp]\label{figure5}
\centering
\subfloat[]{\includegraphics[width = 0.47\columnwidth]{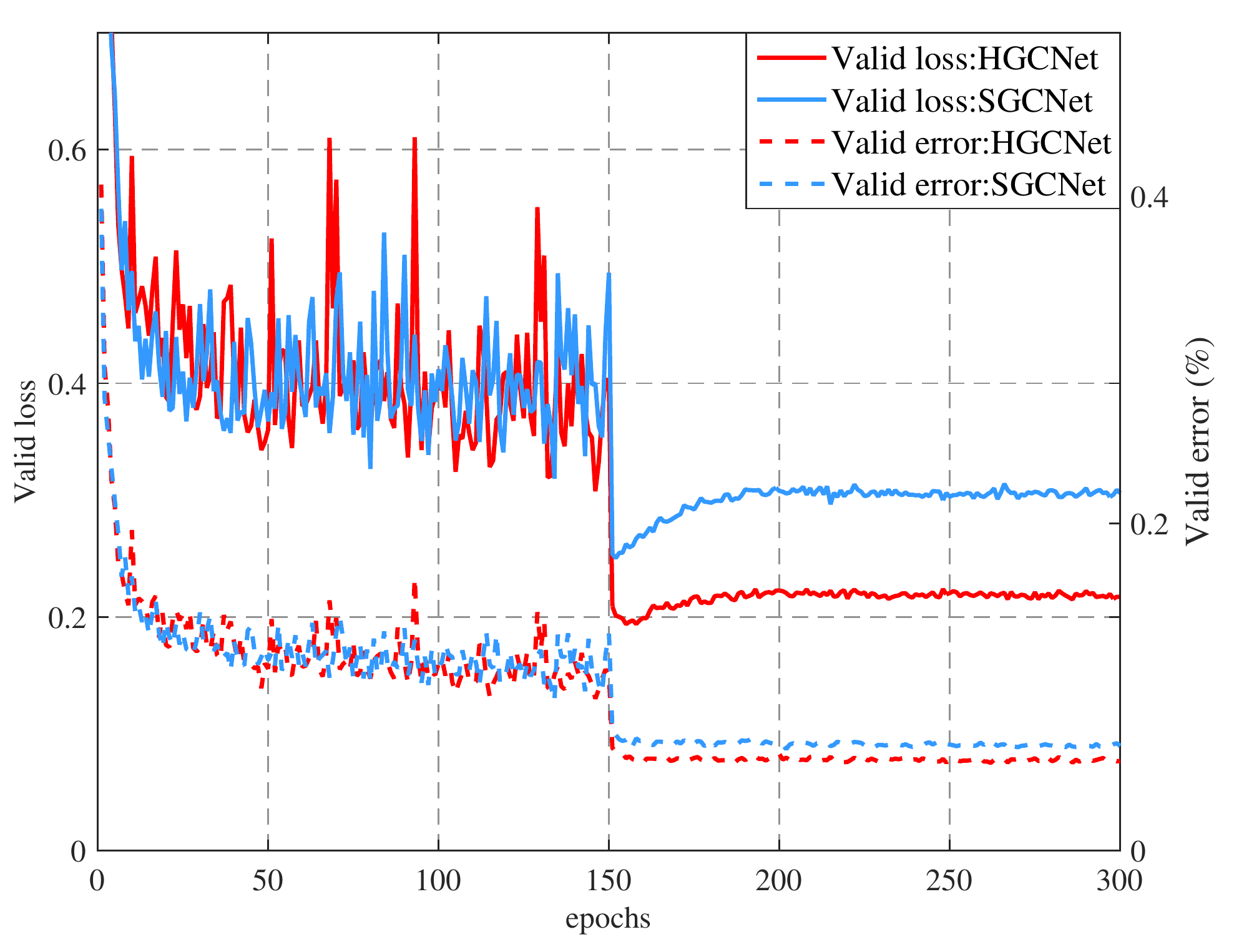}}\
\hspace{0.5cm}
\subfloat[]{\includegraphics[width = 0.47\columnwidth]{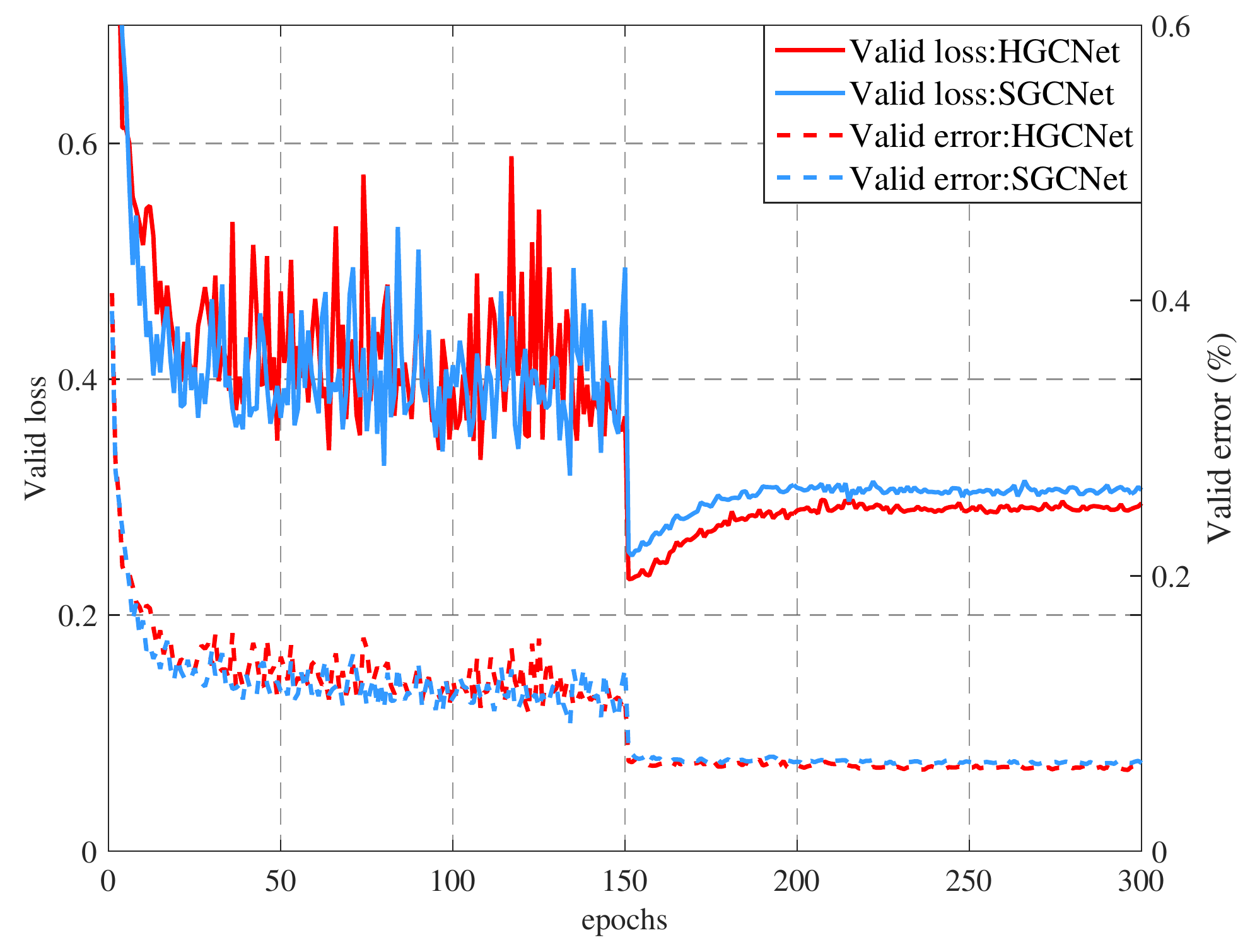}}\
\caption{ Compare the optimization between SGC and the proposed HGC operation. (a) is the valid loss and valid error with the same number of groups, and (b) is under the same number of parameters.}
\end{figure*}
\subsection{Ablation study on CIFAR}
\label{VII}
We first perform a set of experiments on CIFAR-10 to validate the effectiveness of the efficient HGC operation and the proposed HGCNets.

\textbf{Training details}. We train all models with stochastic gradient descent (SGD) using similar optimization hyperparameters as in \cite{4,11}, Specifically, we adopt Nesterov momentum with a momentum weight decay of $10^{-4}$. All models are trained with mini-bath size 128 for 300 epochs, unless otherwise specified. We use a cosine shape learning rate which starts from 0.1 and gradually reduces to 0.

\textbf{Ablation Study}. For better contrast with standard group convolution (SGC), we replace the hierarchical group convolution with SGC in the HGC module which is formed the SGCNets. We first explore the accuracy of them with respect to different number of groups, the results are shown in Table 1 and Figure 4(a). When the group number is kept the same, HGCNets surpass SGCNets by a large margin.  As can be seen, the accuracy drops dramatically when the standard group convolution is applied to the $1\times 1$ convolution, mainly due to the loss of representation capability from hard assignment. Differently, our HGC successfully generates more discriminative features and maintains the accuracy even with large number of groups. More importantly, HGCNets gain substantial improvements as the group number increases. Figure 4(b) shows the computational efficiency gains brought by the HGC. Compared to SGCNets, HGCNets require 30\% fewer parameters to achieve comparable performance.

As discussed above, increasing $G$ makes more inter-group connections lost, which aggravates the loss of inter-group information and harms the representation capability. However, the hierarchical group convolution fuses the features from all channels hierarchically and generates more discriminative features than ShuffleNet. As shown in Figure 5, HGCNet overcomes the performance degradation  and has a better convergence than the network which uses standard group convolution. These improvements are consistent with our initial motivation to design HGC module.
\subsection{Comparison to state-of-the-art compact CNNs}
\label{VIII}
In Table 2, we show the results of experiments comparing HGCNets with alternative state-of-the-art compact CNN architectures. Following \cite{11}, our models were trained for 300 epochs, and set $G$ to 4 for better tradeoff between the compression and accuracy. From the results, we can observe that HGCNets require fewer parameters and FLOPs to achieve a better accuracy than MobileNets and ShuffleNets.
\begin{table}[hbp]
  \caption{Comparison to state-of-the-art compact CNNs}
  \begin{center}
  \begin{tabular}{c c c c c}
  \toprule
    Model  & FLOPs & Params & CIFAR-10 & CIFAR-100 \\
    \midrule
    ShuffleNet\cite{8} & 161M & 0.91M & 7.71 & 29.94 \\
    \midrule
    MobileNetV2\cite{10} & 158M & 1.18M & 6.04 & 31.92\\
    \midrule
    HGCNet-67 & 112M & 0.58M & 5.61 & 25.83\\
    \midrule
    HGCNet-91 & 187M & 0.97M & 5.53 & 24.42\\
  \bottomrule
  \end{tabular}
  \end{center}
\end{table}
\subsection{Comparison to state-of-the-art large CNNs}
\label{IX}
In this subsection, we experimentally demonstrate that the proposed HGCNets, as a lightweight architecture, can still outperform state-of-the-art large models, e.g., ResNet\cite{4}. We can also integrate the SE-block \cite{28} to the HGC module to adaptively recalibrate channel-wise feature responses by explicitly modeling importance of each channel. As shown in Table 3, the original HGCNets can already outperform 110-layer ResNet using 6x fewer parameters.
When we insert SE block into HGC module, the top-1 error of HGCNet on CIFAR-10 further decreases to 5.81\%, with negligible increase in the number of parameters.
\begin{table}[hbp]
  \caption{Comparison to state-of-the-art large CNNs}
  \begin{center}
  \begin{tabular}{c c c c}
  \toprule
    Model  & Depth & Params & CIFAR-10 \\
    \midrule
    VGG\cite{2} & 16 & 5.40M & 6.60 \\
    \midrule
    ResNet\cite{4} & 56 & 0.85M & 6.97 \\
    \midrule
    ResNet\cite{4} & 110 & 1.5M & 6.43 \\
    \midrule
    ResNet\cite{4} & 164 & 1.7M & 5.93 \\
    \midrule
    HGCNet & 42 & 0.28M & 6.36 \\
    \midrule
    SE-HGCNet & 42 & 0.37M & 5.81 \\
  \bottomrule
  \end{tabular}
  \end{center}
\end{table}
\section{Conclusion}
\label{VI}
In this paper, we propose a novel hierarchical group convolution operation to perform model compression by replacing standard group convolution in deep neural networks. Different from standard group convolution which blocks the inter-group information exchange and induce the severe performance degradation, HGC can effectively leverage the inter-group information and generate more discriminative features even with a large number of groups. Based on the proposed HGC, we propose HGCNets, a new family of compact neural networks. Extensive experiments show that HGCNets achieve higher classification accuracy than the prior CNNs designed for mobile devices at the same or even lower complexity.

\bibliographystyle{IEEETrans}
\bibliography{ref}
\end{document}